\documentclass[conference]{IEEEtran}
\IEEEoverridecommandlockouts
\usepackage{cite}
\usepackage{amsmath,amssymb,amsfonts}
\usepackage{algorithm,algorithmic}
\usepackage{graphicx}
\usepackage{textcomp}
\usepackage{epsfig}
\usepackage{amssymb}
\usepackage{amsthm}
\usepackage{mathtools}
\usepackage{lineno}
\def\BibTeX{{\rm B\kern-.05em{\sc i\kern-.025em b}\kern-.08em
    T\kern-.1667em\lower.7ex\hbox{E}\kern-.125emX}}
\begin{document}

\title{Dynamic Approach for Lane Detection using Google Street View and CNN}

\author
{\IEEEauthorblockN{Rama Sai Mamidala}
\IEEEauthorblockA{
\textit{National Institute of Technology}\\
Surathkal, Mangalore, India\\
ramasai840@gmail.com}
\and
\IEEEauthorblockN{Uday Uthkota}
\IEEEauthorblockA{
\textit{National Institute of Technology}\\
Surathkal, Mangalore, India\\
uday7u7@gmail.com}

\and
\IEEEauthorblockN{Mahamkali Bhavani Shankar}
\IEEEauthorblockA{
\textit{National Institute of Technology}\\
Surathkal, Mangalore, India\\
mbshankar97@gmail.com} 

\and \hspace{2.5cm}
\IEEEauthorblockN{A. Joseph Antony}
\IEEEauthorblockA{
\textit{\hspace{2.5cm} Insight Centre for Data Analytics, Dublin City University}\\\hspace{2.5cm}
Dublin, Ireland\\\hspace{2.5cm}
josejaihind@gmail.com}

\and \hspace{2cm}
\IEEEauthorblockN{A. V. Narasimhadhan}
\IEEEauthorblockA{
\textit{\hspace{2cm}National Institute of Technology}\\ \hspace{2cm}
Surathkal, Mangalore, India\\\hspace{2cm}
dhan257@gmail.com}
}

\maketitle

\begin{abstract}
Lane detection algorithms have been the key enablers for a fully-assistive and autonomous navigation systems. In this paper, a novel and pragmatic approach for lane detection is proposed using a convolutional neural network (CNN) model based on SegNet encoder-decoder architecture. The encoder block renders low-resolution feature maps of the input and the decoder block provides pixel-wise classification from the feature maps. The proposed model has been trained over 2000 image data-set and tested against their corresponding ground-truth provided in the data-set for evaluation. To enable real-time navigation, we extend our model's predictions interfacing it with the existing Google APIs evaluating the metrics of the model tuning the hyper-parameters. The novelty of this approach lies in the integration of existing segnet architecture with google APIs. This interface makes it handy for assistive robotic systems. The observed results show that the proposed method is robust under challenging occlusion conditions due to pre-processing involved and gives superior performance when compared to the existing methods.
\end{abstract}


\section{Introduction}
\begin{figure}
	\centering
	\includegraphics[scale =0.45]{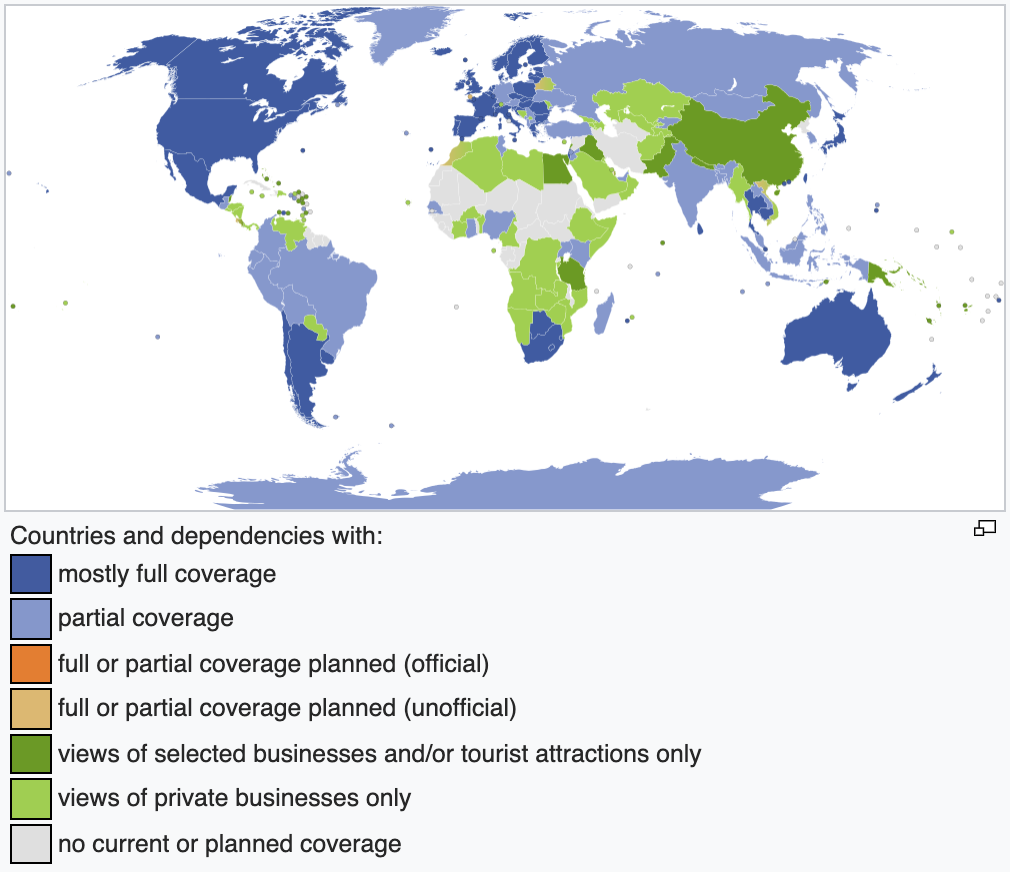}
	\vspace*{\fill}
	\caption{Availability of Google Street View images, till 2018
	\texttt{http://support.google.com}}

	\label{fig:2}
\end{figure}

 Recently most of the high-end cars come with advanced features like semi-autonomous driving systems, vehicle-to-vehicle (V2V) and vehicle-to-infrastructure (V2I) communication, automatic emergency braking, self-parking, night vision and back up cameras. This intervention of machines in human driving system making it more free from traffic hassle and accidents is the underneath motivation. With five of all the six levels of autonomy (L1-L5) defined by the NHTSA (US National Highway Traffic Safety Administration) has the lane detection as the primary requirement \cite{b24} in occluded scenes. Tesla's Autopilot and Mercedes E-class's Drive pilot are some of the best examples for breakthrough in semi-autonomous driving systems. Allowing the car/vehicle to access the location of the system is a foundation for any navigational system.\\
Many of the proposed techniques for lane detection use traditional computer vision and image processing techniques \cite{b7}. They often incorporate highly-specialized and hand-crafted features, but with worst computational and run-time complexities. The introduction of deep neural networks have become very handy to incorporate more number of feature maps for a better efficient model \cite{b7}.
Convolutional Neural Network (CNN) is the primitive deep-net model that has a wide scope of variations in the architecture based on the architectural parameters \cite{b13}. We interface our deep-net models to the existing Google APIs like Google maps, street-view, and geo-locate APIs to access real-time data for navigation. This is one of the major contributions of this paper. The models obtained through deep-nets are quite efficient when trained over a high performance GPU (Graphic processing unit). A cluster-based GPU has been used for training our networks using SLURM management system, as in \cite{b15}.\\
The retrieval of images from Google Street-View is quite a powerful alternative rather than camera-based images in certain applications like drawing out the coverage of plantation in a region or detecting traffic signs/figures on the lane and take steps to improve the awareness. Mainly for the earlier stages of development huge image data-sets are available for many different countries all over the world (\ref{fig:2}) which can be accessed easily and efﬁciently for further developments on them as depicted in this paper.\\

Google API's are integrated with our model to facilitate the real-time data using proper authentication key and certain python libraries (like geolocate and google\_streetview). To demonstrate the efficacy of SegNet, we present a real-time approach of lane segmentation for assistive navigation which can be scaled to more number of classes for autonomous navigation \cite{b17} based on the use-case. The data-set utilized for training the proposed network was created by the researchers from the University of Wollongong, Australia \cite{b1}. We trained our deep net models by tuning the hyper-parameters, and monitoring the learning curves (\ref{fig:3} and \ref{fig:4}) for proper convergence. We evaluated the models in real-time based on the mean-squared between the lane segmented and the lane expected images.

\section {\textbf{Existing methods}}

In the last two decades, there has been a huge interest in automatic lane detection systems \cite{b3, b12, b16}. Though there are a lot of existing methods for lane detection in marked roads \cite{b2} and even for unmarked roads \cite{b4,b5}, they didn't consider the real-time obstacles such as occlusion in the roads, bad weather conditions, and varied intensity images. Before the introduction of deep networks, some of the efficient models with best performance relied on certain machine-learning classifiers like, Gaussian-mixture model (GMM) \cite{b2}, support-vector machine model (SVM) \cite{b6} and multi-layered Perceptron classifier (MLPC) \cite{b7, b20}. The supervised classifiers, SVM and MLPC, both follow back-propagation \cite{b7} approach for the convergence of the learning rate (alpha) and the weights of the hidden layers respectively. Random forest techniques have been used along certain boosting techniques as shown in \cite{b23}, but the CNN's due to their capture and learn more number of features compared to any random-forest algorithm with variant boosting techniques is less preferred compared to CNN as shown in \cite{b20}. Further for the initialization of weights and learning rate, Xavier initialization Eq\eqref{eq:1}is used, which follows an intuitive approach of finding the distribution of the random-weights. Several dynamic approaches \cite{b17, b18} have been introduced depending on the type of application, of which, lane detection using Open-Street-Map (OSM) \cite{b8} is the most-widely used tool to obtain the test image based on the geographical data. But, OSM cannot provide the current location of the system as shown in \cite{b21}, which makes it incapable of obtaining the test image of the current location of the system. This drawback can be made overcome by using certain google APIs by which, the current location of the system can be obtained.

 \begin{figure}
	\centering
	\includegraphics[scale =0.3]{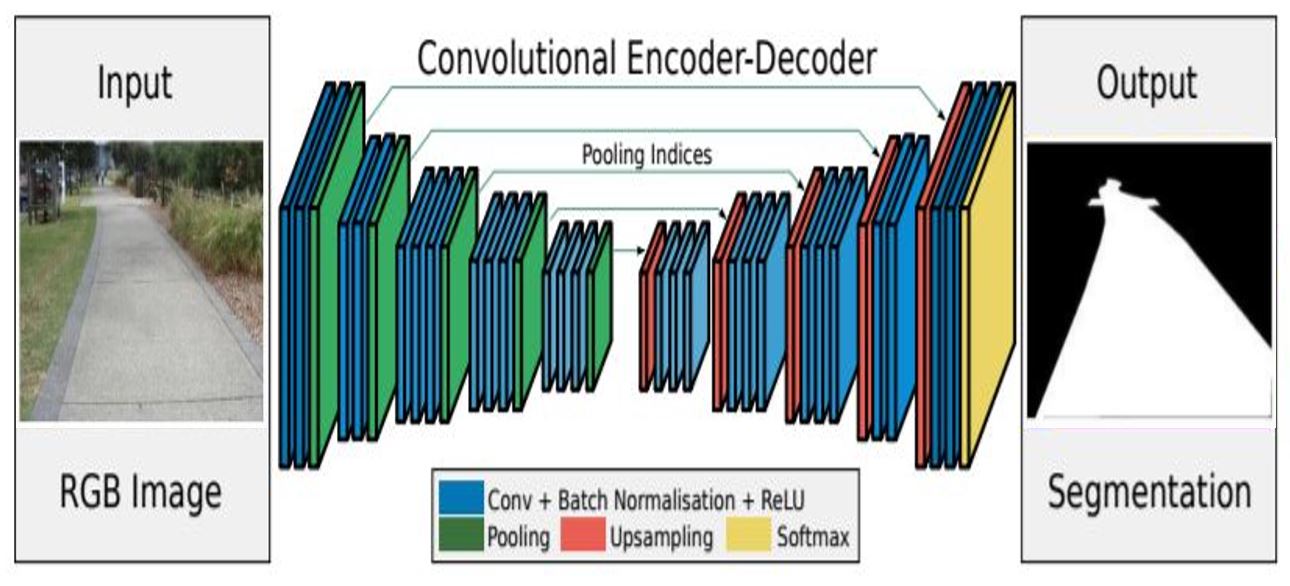}
	\caption{Illustration of Segnet architecture with the input image and the final expected output image.}

	\label{fig:cca}
\end{figure}


\section {\textbf{Proposed Method}}
\subsection{Segnet Model}
Unlike the models \cite{b1, b2, b3, b4} that have been proposed so-far, CNN allow networks to have fewer weights (as these parameters are shared), and they are given a very effective tool - convolutions - for image processing. As with image classification, convolutional neural networks (CNN) works great on semantic-segmentation problems\cite{b9}, even when the background has identical features like the lane. Hence, the current issue of lane segmentation, which comes under semantic segmentation can be solved effectively by using an encoder-decoder type segnet architecture, as shown in Fig.\ref{fig:cca}. The current model has clusters of layers, where each cluster consists of a convolutional layer, activation (rectified linear unit/ ReLu, maps the input matrix with a function that remove the linearity of the model and provides scope to learn more complex functions) and a pooling layer, followed by a de-convolution block (cluster of layers consisting of up-sampling, activation and convolutional layers) after encoding block. The main feature of the SegNet is the use of max-pooling indices in the decoders to perform up-sampling of low resolution feature maps. This retains high frequency details in the segmented images and also reduces the total number of trainable parameters in the decoders. The process of training from scratch gets started by proper initialization of weights, using the proposed Xavier initialization since, no data will be available by default at the beginning of data.\vspace{0.1cm}\\
w$_{i}$ - indicating the weights of the i$^{th}$ kernel\\
b - indicating the bias/threshold of the respective neuron and \\
y - indicating the linear combination of the weight matrix (w$_{i}$) and input tensor(x$_{i}$), which can be represented as follows:

\begin{equation}\label{eq:1}
y = w_{1}x_{1} + w_{2}x_{2} + ....+ w_{n}x_{n} + b
\end{equation}
 Equating the variance along 'x' and 'y', we obtain the following equation:
\begin{equation}\label{eq:2}
    {var(w_{i}) = 1/N}
\end{equation} 

where, 'N' indicates the dimension of image and the weights corresponding to the Gaussian distribution of variance$(\sigma^{2}) = 1/N$ are randomly chosen from the result obtained in Eq\eqref{eq:2}.

\subsection{Training the Model and Tuning the Hyper-parameters}
The model has been trained using dataset created by researchers from University of Wollongong. The dataset contains 2000 RGB images (resolution: $784\times302$). We randomly selected around 1500 images of high variance and the rest utilized for performance evaluation setting up different hyper-parameters. To reduce the time-complexity the size of the image is reduced to $160\times320$ resolution. Since the lane classification comes under binary class, the whole data-set is one-hot encoded \cite{b19} and saved before training the model. The efficiency and the time-complexity of the model depends on the hyper-parameters i.e. epochs and batch-size, where, the batch-size corresponds to the number of images that enter the network at a time and acts as a vital factor to influence the time-complexity; and the number of epochs correspond to the number of times, the whole data-set traverses through the whole network model to avoid under-fitting of the model and enhance the training efficiency of the model. We have trained the network using the open source deep learning framework Keras \cite{b14} on an online Odyssey cluster \cite{b15} based GPU. The proposed model has been trained against 100, 300 and 500 images with the number of epochs 30,30,40 respectively and constant batch-size (20). Finally, the whole data-set (1500 images) with 80 and 115 as the number of epochs and constant batch-size to examine the variations in efficiency and time-complexity with the architectural parameters stored as .JSON and the weights at each layer in a hierarchical data format (HDF5).

\begin{figure}
	\centering
	\includegraphics[scale =0.3]{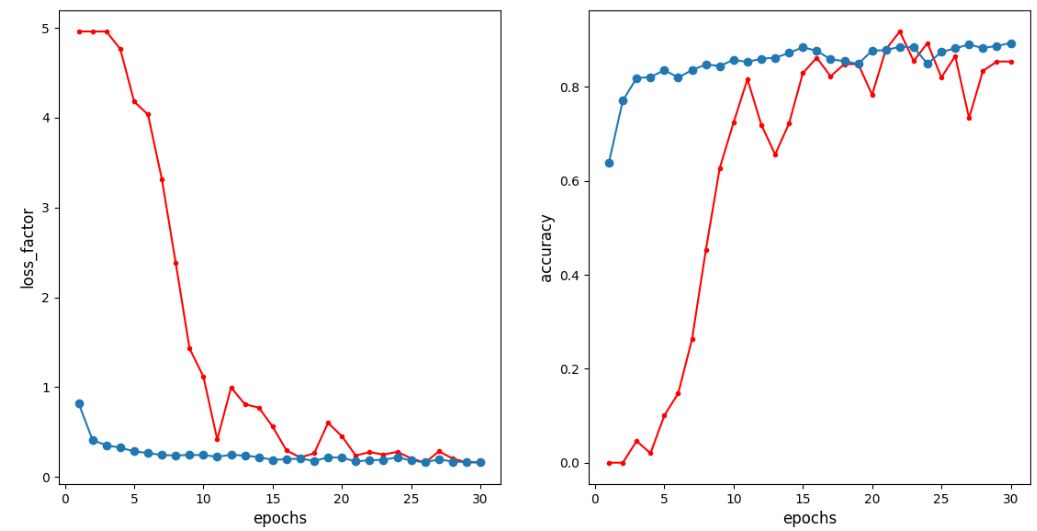}
	\caption{Learning curves (validation accuracy in red and training accuracy in blue) corresponding to the model with 500 training images.}
	\label{fig:3}
\end{figure}

\begin{figure}
	\centering
	\includegraphics[scale =0.3]{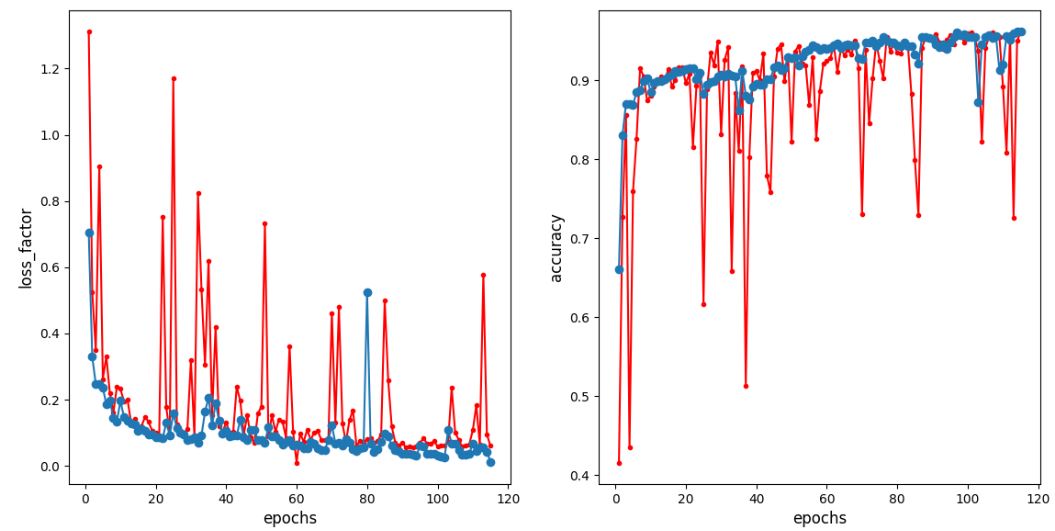}
	\caption{Learning curves (validation accuracy in red and training accuracy in blue) corresponding to the model with 1500 training images.}
	\label{fig:4}
\end{figure}

\subsection{The Dynamic Approach for Segmentation}
\begin{figure}
	\centering
	\includegraphics[width = 7cm, height = 5cm]{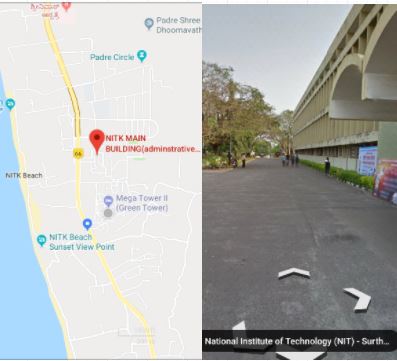}
	\caption{The above figure represents the street-view image (right) obtained, based on the location (left) of the system.}
	\label{fig:5}
\end{figure}

After training the model with the available data-set of 1500 distinct images over GPU, the weights of the trained model are saved and the model is evaluated to get the confusion matrix metrics as shown in Table.\ref{tab:b} and test-accuracy using the loss function (mean-squared error) as given in \eqref{eq:3}.\\
Assistive or autonomous navigation of a robotic system is possible only with the real-time data of the system that can be achieved by the knowledge of location, apart from using camera-based real-time data \cite{b10, b11}. Google API's provide us access the real-time data based on proper authentication key. For the present model, geo-location based lane segmentation can be achieved using OSM API \cite{b14} or Google Maps API, of which, OSM is the most accepted API for it's job. But, the current location of the system, which cannot be obtained using OSM can be achieved using certain Google APIs. Hence, in the current stage, Google geo-locate, Google maps and Google street-view APIs have been used to obtain the street-view image of the system, based on it's geographical location, as shown in Fig.\ref{fig:5}. In this approach, Google geo-locate becomes handy to obtain the current location, followed by Google maps API to obtain it's geographical information, followed by Google Street-View API to obtain the test-image, which is tested on the saved Segnet model architecture to obtain a lane segmented image.

\subsection{Performance Evaluation}
Validation accuracy has been evaluated at each and every epoch to know the training progress of the model, using the following equation:

\begin{equation}\label{eq:3}
    M.S.E = \frac{1}{N} \sum_{j=1}^N \sum_{i=1}^M (f_{i,j}-y_{i,j})^2
\end{equation}
where, N - indicates the count of test data-set \newline M - indicates the dimension of the input kernel and\newline f$_{i,j}$,y$_{i,j}$ - indicate the pixel value of expected and observed images, respectively.\newline
To quantify the accuracy the confusion matrix\cite{b22} was obtained through which true positive (n$_{tp}$), false positive (n$_{fp}$), false negative (n$_{fn}$) and true negative (n$_{tn}$) are extracted to evaluate the precision \eqref{eq:4}, recall \eqref{eq:5}, accuracy \eqref{eq:6} and F$_{1}$-measure \eqref{eq:7} using the following equations: \newline

\begin{equation}\label{eq:4}
    precision = n_{tp}/(n_{tp}+n_{fp})
\end{equation}
\begin{equation}\label{eq:5}
    recall = n_{tp}/(n_{tp}+n_{fn})
\end{equation}
\begin{equation}\label{eq:6}
    accuracy = (n_{tp}+n_{fn})/(n_{tp}+n_{tn}+n_{fp}+n_{fn})
\end{equation}
\begin{equation}\label{eq:7}
    F_{1} = 2\times precision.recall/(precision+recall)
\end{equation}

\begin{table}[t]        
	\caption{Performance of the current Segnet model }\label{tab:a}    
	\centering
	\begin{tabular}{|p{1.2cm}|p{1.2cm}|p{1.2cm}|p{1.2cm}|}
	\hline
		\textbf{Training data-set} & \textbf{Epochs} & \textbf{Testing Data-set}  & \textbf {Test-accuracy}    \\ \hline
		300  & 30 & 100 & 83.4 \% \\ \hline
		500 & 40 & 300 & 85.5 \% \\ \hline
	    1500 & 80 & 500 & 94.7 \% \\ \hline
	    1500 & 115 & 500 & 96.1 \% \\ \hline
	\end{tabular}
\end{table}

\begin{table}[t]        
	\caption{Comparison of metrics across different epochs}\label{tab:b}    
	\centering
	\begin{tabular}{|p{1.2cm}|p{1.2cm}|p{1.2cm}|p{1.2cm}|p{1.47cm}|}
	\hline
		\textbf{Epochs (For 1500 data-set)} & \textbf{Precision} & \textbf{Recall}  & \textbf {Test-accuracy}  &\textbf{$\boldsymbol{F_1}$-measure} \\   \hline
		30 & 0.8990  & \textbf{0.8732} & 0.8340 & 0.9346 \\ \hline
		40 & 0.9334 & 0.8662 & 0.8551 & \textbf{0.9495} \\ \hline
	    80 & 0.9754 & 0.7246 & 0.9470 & 0.9493 \\ \hline
	    115 & \textbf{0.9889} & 0.4042 & \textbf{0.9610} & 0.9445 \\ \hline
	\end{tabular}
\end{table}

\begin{figure*}
	\centering
	\includegraphics[width = 15cm, height = 6cm]{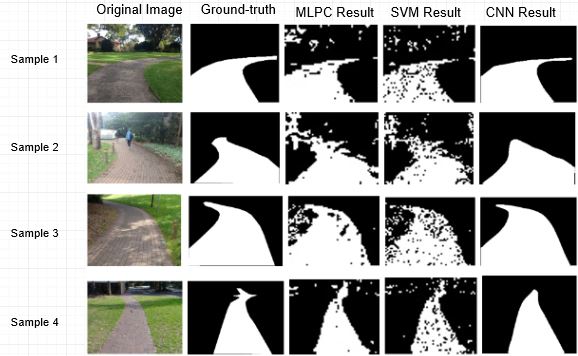}
	\caption{The results of a distinct set of images (col.1) have been listed in above block of figures, with a quite noticeable improvement in the results obtained through CNN (col.5), compared to SVM (col.4) and MLPC (col.3). }
	\label{fig:6}
\end{figure*}

Similarly, certain distinctive test data-sets have been evaluated over different hyper-parameters to obtain the over-all accuracy of the model and the results had been tabulated in Table.\ref{tab:a}. We observed a significant improvement in the results in comparison to the existing methods like SVM, MLPC, as shown in Fig.\ref{fig:6} and a few other adaptive algorithms, as in \cite{b1, b2}.    

\begin{figure*}
	\centering
	\includegraphics[scale = 0.85]{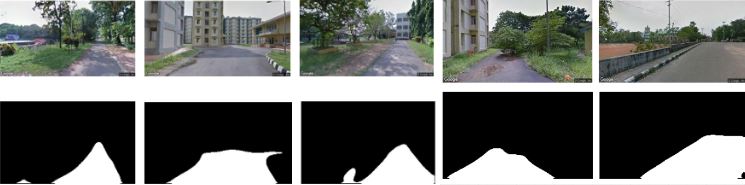}
	\caption{Results representing the dynamic approach, with, Google Street-view images (row -1) in the premises of our institute, and the observed lane-segmented images (row-2).}
	\label{fig:7}
\end{figure*}

\begin{figure*}
	\centering
	\includegraphics[scale =0.8]{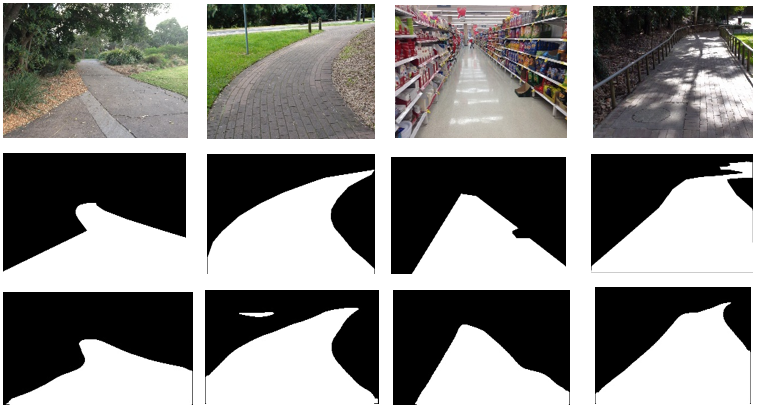}
	\caption{ The above picture represents the results (row -3) obtained using CNN model, of certain exceptional images (row -1), with curves, tiles, indoor-image and image with illumination variation (from left to right in order) along with the corresponding ground-truths (row - 2).}
	\label{fig:8}
\end{figure*}

\section{\textbf{Results and Discussion}}
To start with, we used smaller data-set of 100 images tested on the model trained with 300 images (Table \ref{tab:a}), on a system with specifications of I7 processor, 8GB RAM and 4GB Nvidia G-force GPU. But, the time-complexity hasn't reduced when compared to the previous models. Hence, the model has been implemented on a 8-core cluster based online (cloud) GPU with low time-complexity. The whole data of validation and training accuracies at each epoch, error data if any, along with finals results can be viewed/accessed online using the credentials. With this support, the model has been trained on data-sets of 500 and 1500 images with their labelled ground-truth data, followed by validation on 200 and 300 images respectively (Table \ref{tab:a}), and the test-accuracy is obtained using the rest of data-set of 300 and 500 images respectively. As depicted in (Table \ref{tab:b}), we notice that recall descends with the data-set unlike precision. This is because the classification performed by the model is a pixel-wise classification of occluded images, which is of very high order per image and the reason the saturated recall value for the available data-set is 0.4042 which is not the best for the available data-set due to surplus hardware. At last, both the training and validation accuracies have converged to optimal values (as in Fig.\ref{fig:3} and Fig.\ref{fig:4}). Fig.\ref{fig:6} and Fig.\ref{fig:8} show distinct set of images with occlusions, in these we can observe certain noises (like salt \& pepper) on the results obtained using SVM and MLPC. This is caused by in-efficient training of the model which is due to it's high time-complexity for small data-sets. As tabulated in [Table \ref{tab:a}], effective training with larger data-sets and better efficiency (~97\%) is made possible with CNN's. The images obtained through a set of Google APIs, near the surroundings of our institute (Fig.\ref{fig:7}) have been tested with the correlation coefficient approximated to value 1.

\section{\textbf{Conclusions}}
Most of the recent published work on lane detection is based on marked roads and even the reported work on un-marked-roads do not consider real-time occlusions and are not sufficiently dynamic for real-time implementations. In this paper, SegNet architecture has been compared with other variants to reveal the practical trade-offs involved in designing architectures for segmentation. A dynamic lane detection approach that uses the real-time street images based on the geo-location of the system through Google Street-View API has been proposed in this paper. This approach of interfacing our deep learning model with Google Street-View API is more advantageous compared to OSM, due to it's ability to obtain current location of the system. The current proposed model has the scope of extension to future driver-assistance systems when trained with multiple number of classes for effective lane and obstacle segmentation. By using random-feature set and training data and evaluating different models and converging them to a single stable/adaptive model can be made possible when trained over GPU's like Nvidia TITAN V/X. Also, this can be extended to build an electronic cane for blind people using the data obtained from camera attached to the cane.

\end{document}